\documentclass{llncs}

\usepackage{makeidx}  
\usepackage{amsfonts, amsmath, amsfonts, amssymb}
\usepackage{graphicx} %

\begin{document}
\frontmatter          
\pagestyle{headings}

\title{Combinatorial Ricci Curvature and Laplacians for Image Processing}
\titlerunning{Combinatorial Ricci}  
%
\author{Emil Saucan \and Eli Appleboim \and
Gershon Wolansky  \and
Yehoshua~Y.~Zeevi}
\authorrunning{Emil Saucan et al.}   
%
\tocauthor{Emil Saucan, Eli Appleboim, Gershon Wolansky, Yehoshua~Y.~Zeevi}
\institute{ Technion - Israel Institute of Technology, Haifa 32000, Israel,\\
\email{semil@tx.technion.ac.il, eliap@ee.technion.ac.il, gershonw@math.technion.ac.il, zeevi@ee.technion.ac.il}}

\maketitle              

\begin{abstract}
A new Combinatorial Ricci curvature and Laplacian operators for
grayscale images are  introduced and tested on $2D$ synthetic,
natural and medical images. Analogue formulae for voxels are also
obtained. These notions are based upon more general concepts
developed by R. Forman. Further applications, 
in particular a fitting Ricci flow, are discussed.  
\end{abstract}
\section{Introduction} \label{saucan-sec:intro}
Curvature analysis plays a major role in Image Processing, Computer
Graphics, Computer Vision and their related fields, for many
applications, such as reconstruction, segmentation and recognition,
to list only a few (see, e.g. 
\cite{DLYG}, \cite{LL},
\cite{SAZ}, \cite{YL}). Traditionally, the curvature estimation is
that of  a polygonal (polyhedral) mesh, approximating the ideally
smooth ($\mathcal{C}^2$) image under study, such that the curvature
measures of the mesh converge to the classical, differential,
curvature measure
of the investigated surface. For surfaces, by far the most important curvature is the {\em intrinsic} Gaussian (or total) curvature. 

Recently, partly as an offshoot of the great interest generated by
G. Perelman's  important contribution 
on the Ricci flow and its application in
the proof of Thurston's Geometrization Conjecture, and, implicitly
of the Poincar\'{e} Conjecture (see, e.g. \cite{MT} for a
comprehensive exposition), a flourishing of the study of various
discrete versions of the Ricci flow (and similar related flows)
occurred (see \cite{CL}, \cite{Gl}, \cite{Gu}, \cite{FL}).

Ricci curvature measures the defect of the manifold from being locally Euclidean 
in various tangential directions. More precisely, it appears in the second term of the
formula for the $(n-1)$-volume $\Omega(\varepsilon)$ generated
within a solid angle (i.e. it controls the growth of measured
angles) -- see Fig. 1.
\begin{figure}[h] \label{saucan-fig:Ricci}
\begin{center}
\includegraphics[scale=0.45]{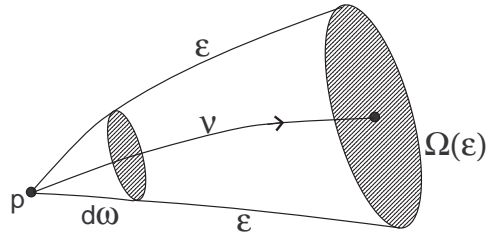}
\end{center}
\caption{Ricci curvature as defect of the manifold from being locally Euclidean 
in various tangential directions (after Berger \cite{Be1}). Here
$d\omega$ denotes the $n$-dimensional solid angle in the direction
of the vector ${\bf v}$, and $\Omega(\epsilon)$ the $(n-1)$-volume
generated by geodesics of length $\varepsilon$ in $d\omega$.}
\end{figure}
%
%
Moreover, 
\[ {\bf v}\cdot Ricci({\bf v}) = \frac{n - 1}{vol\big(\mathbb{S}^{n-2}\big)}\int_{{\bf w} \in T_p(M^n),\; {\bf w}
\perp {\bf v}}\!\!\!\!\!K(<{\bf v},{\bf w}>)\,,
\]
where $<{\bf v},{\bf w}>$ denote the plane spanned by  ${\bf v}$ and
${\bf w}$, i.e. Ricci curvature represents an average of sectional
curvatures. The analogy with mean curvature is further emphasized by
the following remark: Ricci curvature behaves as the Laplacian of
the metric $g$ (see, e.g. \cite{Be}).
It is also important to note that in dimension $n=2$, that is in the
case that is the most relevant for classical Image Processing and
its related fields, Ricci curvature reduces to sectional (and
scalar!) curvature, i.e. to the classical Gauss curvature.

However, both in the more classical context, as well as in the new
directions mentioned above, smooth surfaces and/or their polygonal
approximations considered. Unfortunately, smooth surfaces are at
best a crude model (and usually nothing but a polite fiction), as
far as digital and grayscale images, i.e. the standard objects of
study in Image Processing,  are considered, in particular for those
objects that are not ``natural images'', such as images produced
using ultrasound imaging, MRI or CT.
It would be of practical interest  to define a proper notion of
curvature for digital objects, in the spirit of \cite{H}. By
``proper'' we mean discrete, intrinsic to the nature of the spaces
under investigation, and not an approximation or rough
discretization of a differential notion. Moreover, we hope that, by doing this, we shall be able to help bridge, at least partially, the divide between ``Digital'' and ``Smooth'' Image Processing. In addition, since,as we shall show, each dimension displays its own Laplacian, we believe our method can produce more types of heat flow, along edges, pixels, etc, thus allowing for better tools for the intelligence of higher dimensional data (such as RGB images, images with texture, etc.).

We are fortunate in our quest to be able to rely on the work of R.
Forman \cite{Fo} on Combinatorial Ricci curvature and the so called
``Bochner Method'', where he addressed this very problem in the far
more general setting of weighted {\it cell complexes}, 
 which represent an abstractization of both
polygonal meshes and weighted graphs. While we succinctly present
some of the more general facts residing in Forman's work, in this
paper we concentrate solely on the case of grayscale images with
their very special combinatorics and weights, and largely defer the study of
higher dimensional (color) images and their curvatures and Laplacians for
further study \cite{us}.

The present paper represents an extension of our previous work \cite{miccai}.
More precisely, we have augmented the previous version by adding: (a) The formula for the Ricci curvature of 3-dimensional cubical complexes (that is, in the case of images, for voxels); (b) A discussion on the purely combinatorial version of the Ricci curvature and Laplacians; (c) A theoretical comparison of the classical and combinatorial Laplacians; (d) A combinatorial diffusion, corresponding to upsampling and downsampling (this aspect being augmented by a computational example); (e) A discussion of a possible Ricci flow for our combinatorial setting, with a concrete suggested direction of study.
In addition we have illustrated the meaning of Ricci curvature and included
more detailed exposition of the Algebraic Topology background and ideas.
And, last, but not least, we bring more (and new) experimental results: on synthetic images (that were not given previously), on standard test images and also on medical images.
%
%
\section{Forman's Combinatorial Ricci Curvature} \label{saucan-sec-FormanCurv}
We sketch below some of Forman's main ideas \cite{Fo}. 
This requires some technical 
definitions and notations. 
We do not introduce here the
basic technical notions in Algebraic Topology and Differential
Geometry, and refer the reader to \cite{RS} for the former and to
\cite{Be} for the later.

To generalize the notion of Ricci curvature, in a manner that would include weighted cell complexes, one starts
from the following form of the {\it Bochner-Weitzenb\"{o}ck} formula (see, e.g. \cite{Be}) for the {\it
Riemann-Laplace operator} $\Box_p$ on $p$-forms on (compact) Riemannian manifolds:
\begin{equation} \label{saucan-eqn:1}
\Box_p = dd^* + d^*d = \nabla_p^*\nabla_p  + Curv(R)\,,
\end{equation}
where $\nabla_p^*\nabla_p$ is the {\it Bochner} (or {\it rough}) {\it Laplacian}  and where $Curv(R)$ is an 
expression with linear coefficients of the {\it curvature tensor} 
(Here $\nabla_p$ is the {\it covariant derivative} operator.) 
%
%
Of course, for cell-complexes one cannot expect such differentiable
operators. However, a {\it formal} differential exists: in our
combinatorial context (the operator) ``$d$'' being replaced by
``$\partial$'' -- the boundary operator of the cellular chain
complex (see \cite{RS}),
\[0 \longrightarrow C_n(M,\mathbb{R}) \stackrel{\partial}{\longrightarrow} C_{n-1}(M,\mathbb{R}) \stackrel{\partial}{\longrightarrow}
\cdots \stackrel{\partial}{\longrightarrow} C_{0}(M,\mathbb{R}) \longrightarrow 0\,,\]
were cells are playing in this setting the role of the forms in the classical (i.e. Riemannian) one.
The following definition of the combinatorial Laplacian becomes now
natural:
\begin{equation} \label{saucan-eqn:2}
\Box_p = \partial\partial^* + \partial^*\partial: C_p(M,\mathbb{R}) \rightarrow C_p(M,\mathbb{R})\,,
\end{equation}
%
%
where  
$\partial^*:C_p(M,\mathbb{R}) \rightarrow C_{p+1}(M,\mathbb{R})$ is the {\it adjoint} (or {\it coboundary})
operator of $\partial$, defined by:
\(<\partial_{p+1}c_{p+1},c_p>$ $=
<c_{p+1},\partial_p^*c_p>_{p+1}\,,\)
where $<\cdot,\cdot> = <\cdot,\cdot>_p$ is a (positive definite)
inner product on $C_p(M,\mathbb{R})$, i.e. satisfying: (i)
$<\alpha,\beta> = 0, \forall \alpha \neq \beta$ and (ii)
$<\alpha,\alpha> = w_\alpha > 0$ -- the weight of cell $\alpha$.
%
%

 Forman \cite{Fo} shows that an analogue of the Bochner-Weitzenb\"{o}ck formula holds in this setting, i.e. that
there exists a canonical decomposition of the form:
\begin{equation} \label{saucan-eqn:3}
\Box_p = B_p + F_p\,,
\end{equation}
where $B_p$ is a {\it non-negative operator} and $F_p$ is a certain diagonal matrix. $B_p$ and $F_p$ are called,
in analogy with the classical Bochner-Weitzenb\"{o}ck formula, the {\it combinatorial Laplacian} and {\it
combinatorial curvature function}, respectively.
Moreover, if $\alpha = \alpha^p$ is a $p$-dimensional cell (or $p$-cell, for short), then we can define the {\it
curvature function}:

\begin{equation} \label{saucan-eqn:4}
\mathcal{F}_p = <F_p(\alpha),\alpha>,
\end{equation}
$F_p: C_p \rightarrow C_p$\, being regarded as a linear function on $p$-chains.
For dimension $p=1$ we obtain, by analogy with classical case, the
following definition of discrete (weighted) {\it Ricci curvature}:

\begin{definition}  \label{saucan-def:Ricci}
Let $\alpha = \alpha^1$ be a 1-cell (i.e. an edge). Then the {\em Ricci curvature} of $\alpha$ is defined as:
\begin{equation}
{\rm Ric}(\alpha)= \mathcal{F}_1(\alpha).
\end{equation}
\end{definition}
While general weights are possible, making the combinatorial Ricci curvature extremely versatile, it turns out
(see \cite{Fo}), 
that it is possible to restrict oneself only to so called {\it
standard weights}:

\begin{definition}  \label{saucan-def:st-wgh}
The set of weights $\{w_\alpha\}$ is called a {\em standard set of
weights} iff there exist $w_1, w_2 > 0$ such that given a $p$-cell
$\alpha^p$, the following holds:
\[w(\alpha^p) = w_1\cdot w_2^p\,.\]
\end{definition}
(Note that the combinatorial weights $w_\alpha \equiv 1$ represent a
set of standard weights, with $w_1 = w_2 = 1$.)
Using standard weights, we obtain the following formula for polyhedral (and in fact much more general) complexes:
\begin{equation} \label{eq:Forman}
\hspace*{-2cm} \mathcal{F}(\alpha^p) =
w(\alpha^p)\left[\left(\sum_{\beta^{p+1}>\alpha^p}\frac{w(\alpha^p)}{w(\beta^{p+1})}
+
\sum_{\gamma^{p-1}<e_2}\frac{w(\gamma^{p-1})}{w(\alpha^p)}\right)\right.
\end{equation}
\[ 
\left.
- \sum_{\alpha_1^p\parallel \alpha^p, \alpha_1^p \neq
\alpha^p}\left|\sum_{\beta^{p+1}>\alpha_1^p,\beta^{p+1}>\alpha^p}\frac{\sqrt{w(\alpha^p)w(\alpha_1^p)}}{w(\beta^{p+1})}
-
\sum_{\gamma^{p-1}<\alpha_1^p,\gamma^{p-1}<\alpha^p}\frac{w(\gamma^{p-1})}{\sqrt{w(\alpha^p)w(\alpha_1^p)}}\right|\right],
\]
where $\alpha < \beta$ means that $\alpha$ is a face of $\beta$, and
the notation $\alpha_1 \parallel \alpha_2$ signifies that the
simplices $\alpha_1$ and $\alpha_2$ are {\it parallel}, the notion
of parallelism being defined as follows:
\begin{definition}  \label{saucan-def:parallel}
Let $\alpha_1 = \alpha_1^p$ and $\alpha_2 = \alpha_2^p$ be two
p-cells. $\alpha_1$ and $\alpha_2$ are said to be  {\em parallel}
($\alpha_1
\parallel \alpha_2$) iff either:
 (i) there exists $\beta = \beta^{p+1}$, such that $\alpha_1, \alpha_2 <
\beta$; or (ii) there exists $\gamma = \beta^{p-1}$, such that
$\alpha_1, \alpha_2 > \gamma$ holds, but not both simultaneously. (For example, in Fig. 1, $e_1,e_2,e_3,e_4$ are all the edges parallel to $e_0$.)
\end{definition}

Together with the formula above, the (dual) formula for the combinatorial Laplacian (see \cite{Fo}) is also
obtained to be:
\begin{equation}  \label{saucan-eqn:Lap}
\Box_p(\alpha_1^p,\alpha_2^p) =
\sum_{\beta^{p+1}>\alpha_1^p,\beta^{p+1}>\alpha_2^p}\epsilon_{\alpha_1,\alpha_2,\beta}\frac{\sqrt{w(\alpha_1^p)w(\alpha_2^p)}}{w(\beta^{p+1})}
\end{equation}
\[\hspace*{2cm} + \sum_{\gamma^{p-1}<\alpha_1^p,\gamma^{p-1}<\alpha_2^p}\epsilon_{\alpha_1,\alpha_2,\gamma}\frac{w(\gamma^{p-1})}{\sqrt{w(\alpha_1^p)w(\alpha_2^p)}}\,,\]
where $\epsilon_{\alpha_1,\alpha_2,\beta},\epsilon_{\alpha_1,\alpha_2,\gamma} \in \{-1,+1\}$ depend on the
relative orientations of the cells.

\section{Combinatorial Ricci Curvature of Images} \label{saucan-sec:Images}
Before developing the relevant formulae in the special combinatorial
setting of the tilling by squares of the plane, as it is usually
considered in (Discrete) Image Processing, let us first indicate that it is advantageous to use standard weights.
Such natural weights are proportional to the geometric content (s.a. length and area). It follows that the weight of any vertex is
$w(v) = 0$. Bearing this in mind, and considering the combinatorics of the square tilling (see Fig. 1), the specific form of Combinatorial Ricci curvature for $2D$ images is: 
 %
%
%
\begin{equation} \label{eq:Ricci-Forman2D}
\hspace*{-0.5cm}{\rm Ric}(e_0) =
w(e_0)\left[\left(\frac{w(e_0)}{w(c_1)} +
\frac{w(e_0)}{w(c_2)}\right) -
\left(\frac{\sqrt{w(e_0)w(e_1)}}{w(c_1)} +
\frac{\sqrt{w(e_0)w(e_2)}}{w(c_2)}\right)\right]\,.
\end{equation}

For the Laplacian there exists more than one possible choice, depending on the dimension $p$. 
The simplest, and operating on cells of the same dimensionality as
the Discrete Ricci curvature, is  $\Box_1$. Because vertices have
weight $0$ and adjacent cells have opposite orientations, Equation
(7) becomes, in this case (using the notation of Fig. 1):
\begin{equation} \label{saucan-eqn:box1}
\Box_1(e_0) = \Box_1(e_0,e_0) = 
 \frac{w(e_0)}{w(c_1)} -  \frac{w(e_0)}{w(c_2)}.
\end{equation}

The formula for the {\it Combinatorial Bochner Laplacian} follows immediately:
\begin{equation} \label{saucan-eqn:Boch1}
B_1(e_0) = \Box_1(e_0) - {\rm Ric}(e_0)\,.
\end{equation}

Instead of computing a Laplacian {\it along} the edge $e_0$, one can
compute a Laplacian operating {\it across} the edge, namely
$\Box_2(c_1,c_2)$. Since no 3-dimensional cells exist, the first sum
in Equation (7) vanishes. Hence, we have (up to sign):
\begin{equation} \label{saucan-eqn:box2}
\Box_2(c_1,c_2) = \frac{w(e_0)}{\sqrt{w(c_1)w(c_2)}}\,.
\end{equation}
%
%
%
%
%
%
\begin{figure}[h] \label{fig:pixels}
\begin{center}
\includegraphics[scale=0.45]{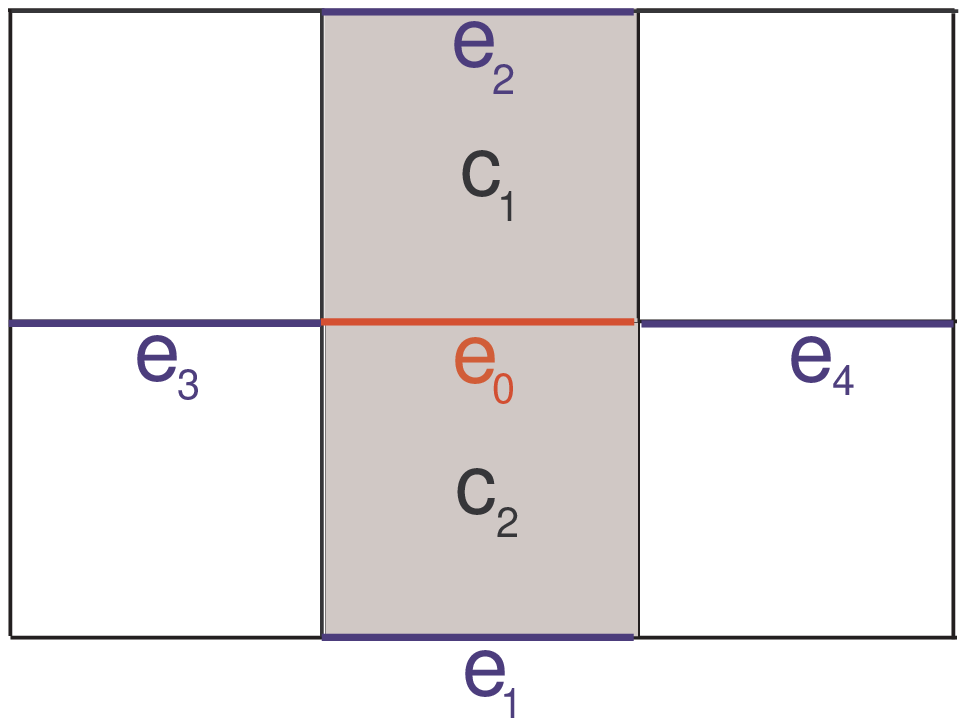}
\includegraphics[scale=0.45]{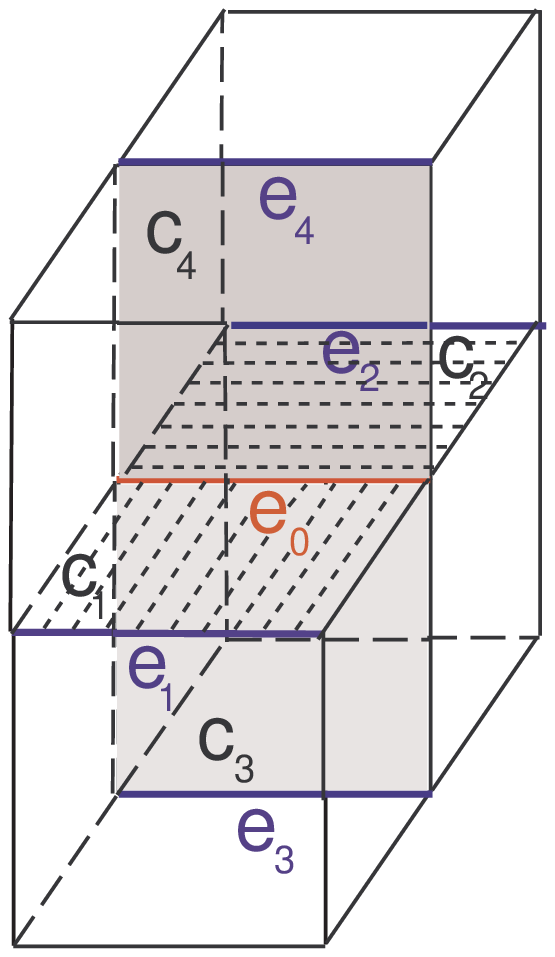}
\end{center}
\caption{The elements appearing in the computation of the Combinatorial Ricci curvature of edge $e_0$ in the
geometry of pixels.}
\end{figure}
%
%
%
\begin{remark}
By ``weighing'' the combinatorial formula for the curvature function $\mathcal{F}_p$ (cf, \cite{Fo}) 
we obtain:
\begin{equation}  \label{saucan-eqn:comb}
\mathcal{F}_p = \sharp\{\beta^{p+1} > \alpha\} + \sharp\{\gamma^{p-1} < \alpha\} - \sharp\{\delta\,|\,\delta \|
\alpha\}.
\end{equation}
(Here $\sharp X$ denotes the number of elements of the set $X$.) For $p= dim\,\alpha = 1$, a simplified, purely combinatorial version of Formula
(\ref{eq:Ricci-Forman2D}) is also obtained:
\begin{equation} \label{saucan-eqn:comb1}
{\rm Ric}(e_0) = w(c_1) + w(c_2) - w(e_1) - w(e_2) - w(e_3) - w(e_4) + 2.
\end{equation}
%
However, the above combinatorial version of the Ricci curvature does not provide considerable geometric insight and, therefore, does not yield interesting results.
\end{remark}

In the case of the cubical geometric configuration present
in Digital Images,
Formula \ref{eq:Forman} becomes:
\[\hspace*{-2cm}
{\rm Ric}(e_0) =
w(e_0)\left[\left(\sum_{c^2>e_0}\frac{w(e_0)}{w(c^2)} +
\sum_{c^0<e_2}\frac{w(c^0)}{w(e_0)}\right)\right.\]
\[\hspace*{2.6cm} \left.
- \sum_{e\parallel e_0, e \neq
e_0}\left|\sum_{c^2>e,c^2>e_0}\frac{\sqrt{w(e_0)w(e)}}{w(c^2)} -
\sum_{c^0<e,c^0<e_0}\frac{w(c^0)}{\sqrt{w(e_0)w(e)}}\right|\right].
\]
Since, as we have already noted, for digital images the vertices' weights are always $0$, we obtain the following
expression for ${\rm Ric}(e_0)$:
\begin{equation} \label{saucan-eqn:RicIm}
{\rm Ric}(e_0) =
w(e_0)\left[w(e_0)\left(\sum_{1}^{4}\frac{1}{w(c_i)}\right) -
\sqrt{w(e_0)}\left(\sum_{1}^{4}\frac{\sqrt{w(e_i)}}{w(c_i)}\right)\right],
\end{equation}
(see Fig. 2, right).

%

\section{Experimental Results} \label{saucan-sec:Res}

Before commencing any experiments with the combinatorial Ricci curvature in the context of images, we had to
choose a set of weights for the 2-, 1- and 0-dimensional cells of an image, that is for squares (pixels), their
common edges and the vertices of the tilling of the image by the pixels. Any such choice should, obviously, be as
natural and expressive  as possible for image analysis. The choice of weights was motivated by two factors: the
context of Image Processing, where a natural choice for weights imposes itself (see below) and the desire (and,
indeed, sufficiency, see Section 2) to employ solely standard  weights.

Since natural weights have to be proportional to the dimension of
the cell, it follows immediately that the weight of any vertex
(0-cell) has to be 0. Moreover, in the beginning, it is natural to
choose $w_1 = 1$ and $w_2 = {\rm length\; of\; a\; cell}$. A
somewhat less arbitrary choice for the length (i.e. basic weight) of
an edge, would be ${\rm Length(e) = (dimension\; of\; the\;
picture)^{-1}}$, hence that for the area (i.e. basic weight) of a
pixel $\alpha$ being ${\rm Area(\alpha)  = (dimension\; of\; the\;
picture)^{-2}}$. The proper weight for a cell $\alpha$ should,
however, take into account the gray-scale level (or height)
$h_\alpha$ of the pixel in question, i.e. $w_\alpha = h_\alpha \cdot
{\rm Area(\alpha)}$. This will become, so we hope, clearer in the
following paragraph. The natural weight for an edge $e$ common to
the pixels $\alpha$ and $\beta$ is $|h_\alpha - h_\beta|$. (A less
``purely'' combinatorial choice of cells and weights is discussed in
\cite{us}.)
%
%

Note that, given an edge $e$, the Ricci curvature ${\rm Ric}(e)$
represents in a way a generalized mean of the weights the cells {\it
parallel} to $e$. 
Therefore, it represents a measure of flow in the direction {\it
transversal} to $e$. It follows, that, contrary perhaps to
intuition, this type of Ricci curvature (and the Bochner Laplacian
associated to it) in direction, say, parallel to the $x$-axis, is
suitable for the
detection of edges and ridges in the $y$-direction. 
 On the other hand, since scalar (i.e. Gauss) curvature, is
 associated to each pixel, that is to each square of the
 tessellation, to compute the Gaussian curvature one has to compute
 the arithmetic mean of the Ricci curvatures of edges of the square
 under consideration -- see Fig. 2. (A similar argument holds if one wishes to
 compute the 1-Laplacians, $\Box_1$ and $B_1$, of a given pixel.)
 The difference between the Ricci curvature computed in the horizontal
 and vertical directions, as well the ``true'', i.e. average Ricci
 curvature can be seen in Fig. 3 and Fig. 4.
 \begin{figure}[h] \label{saucan-fig:synthetic}
\begin{center}
\includegraphics[scale=0.38]{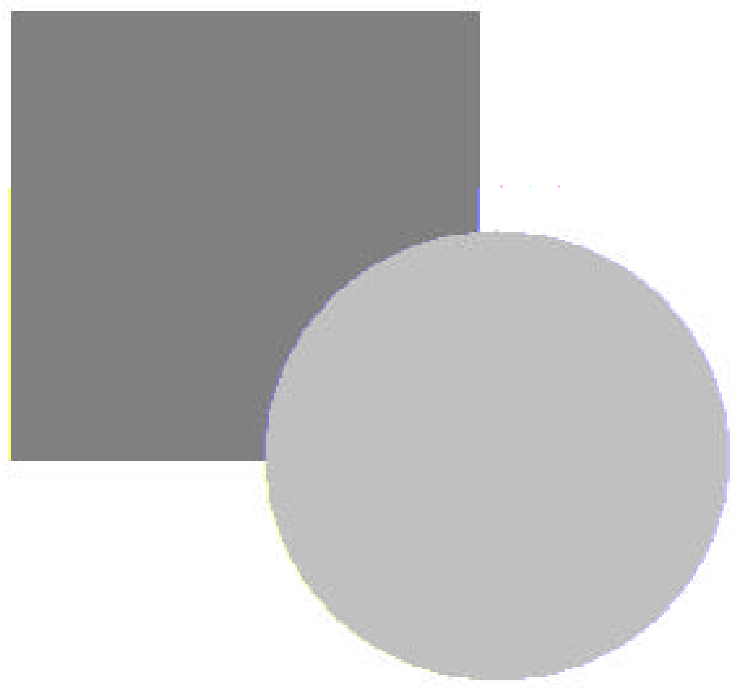}
\includegraphics[scale=0.53]{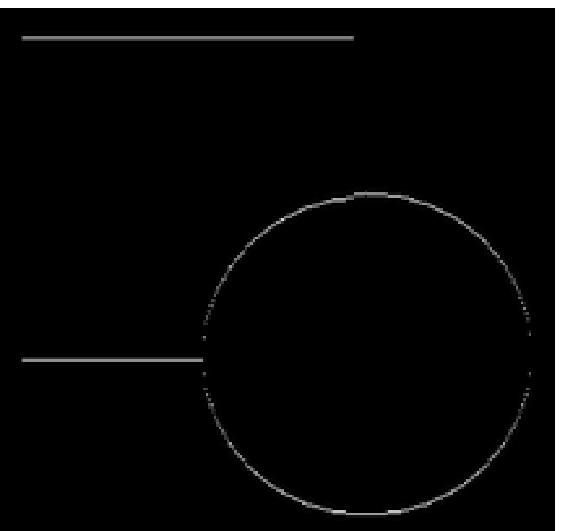}
\includegraphics[scale=0.35]{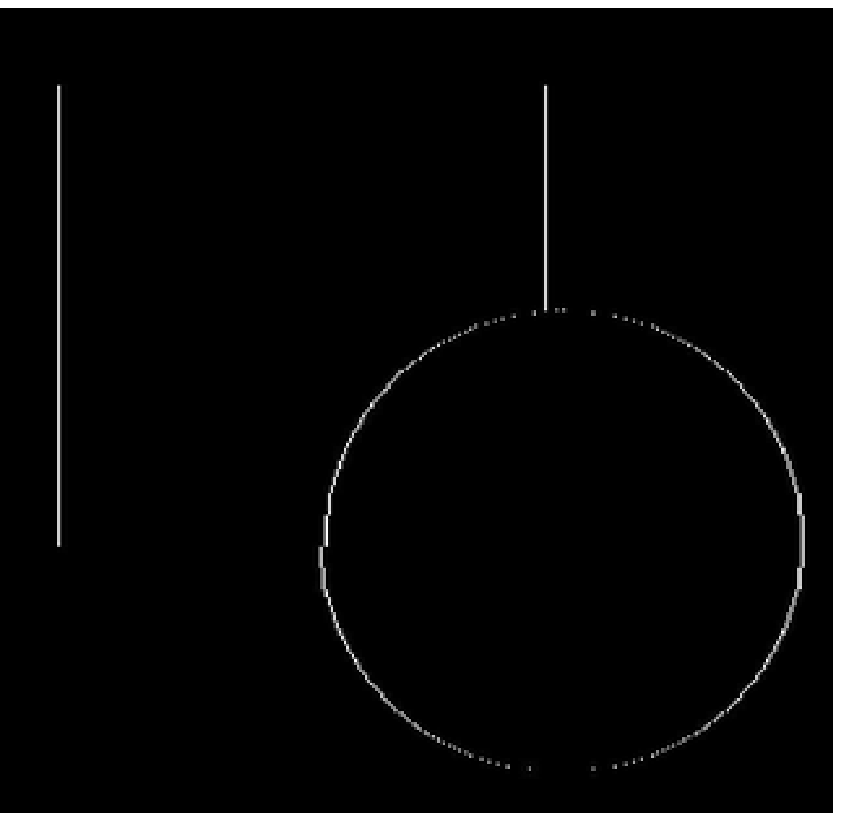}
%
%
\end{center}
\caption{Computing the Ricci curvature for a pixel: Synthetic test
image (left), Ricci curvature computed in horizontal (middle) and
vertical (right) directions.}
\end{figure}
 \begin{figure}[h] \label{saucan-fig:cameraman}
\begin{center}
\includegraphics[scale=0.6]{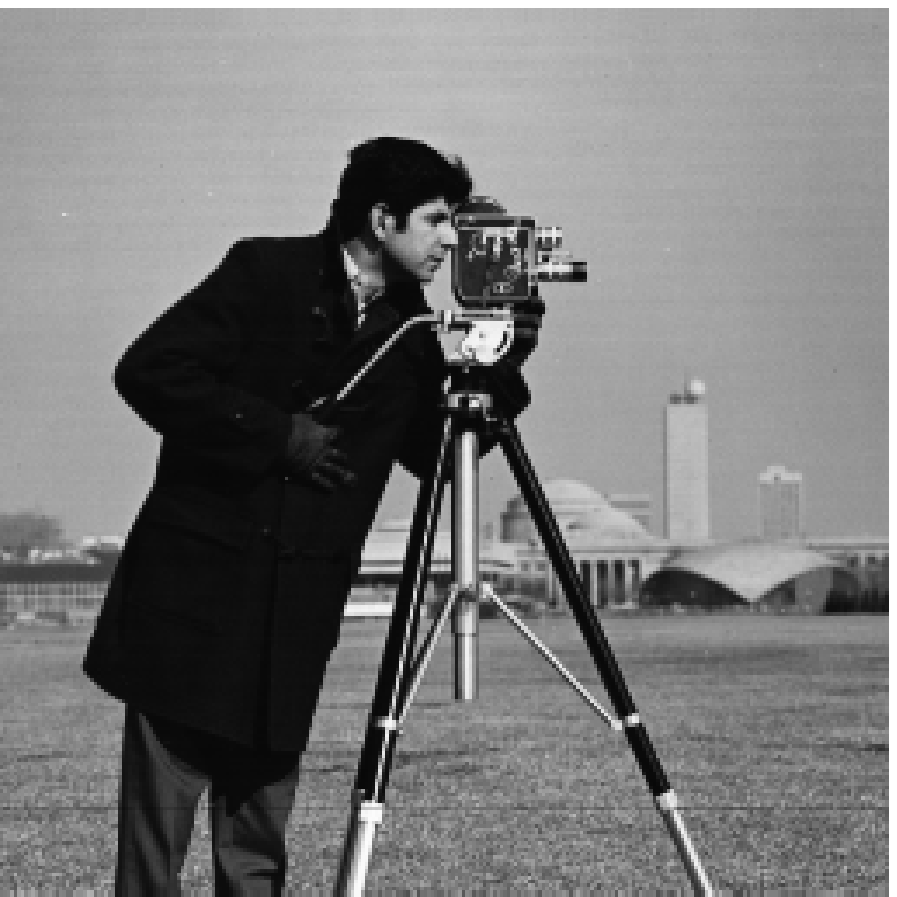}
\includegraphics[scale=0.6]{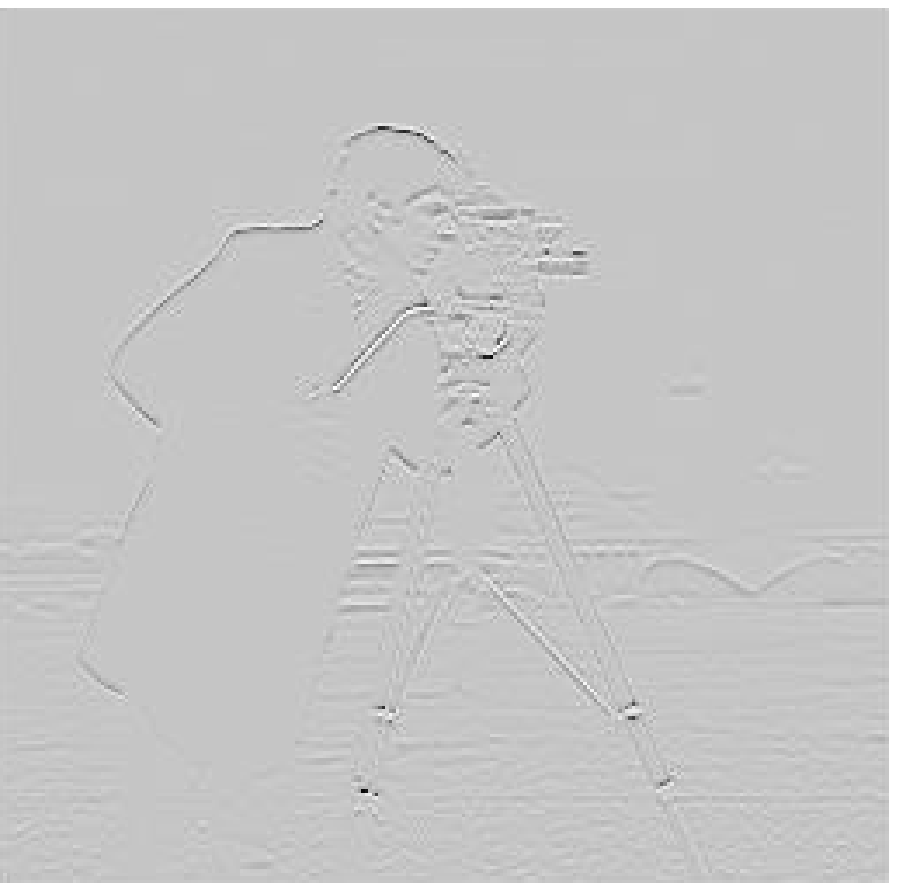}
\includegraphics[scale=0.6]{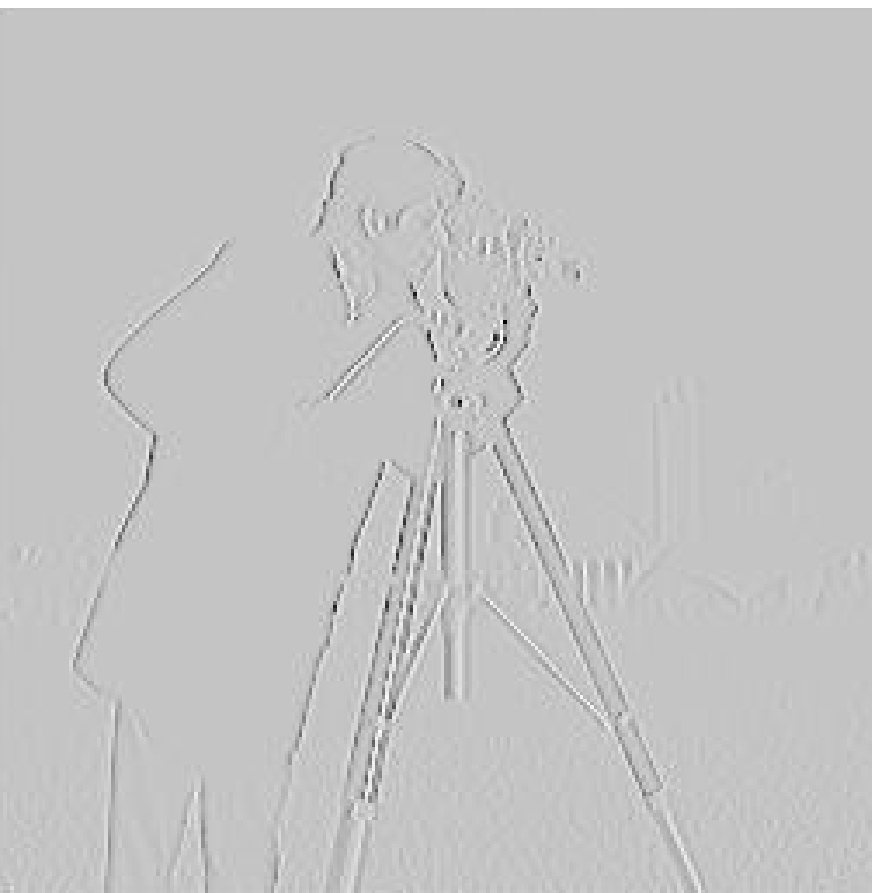}
\includegraphics[scale=0.6]{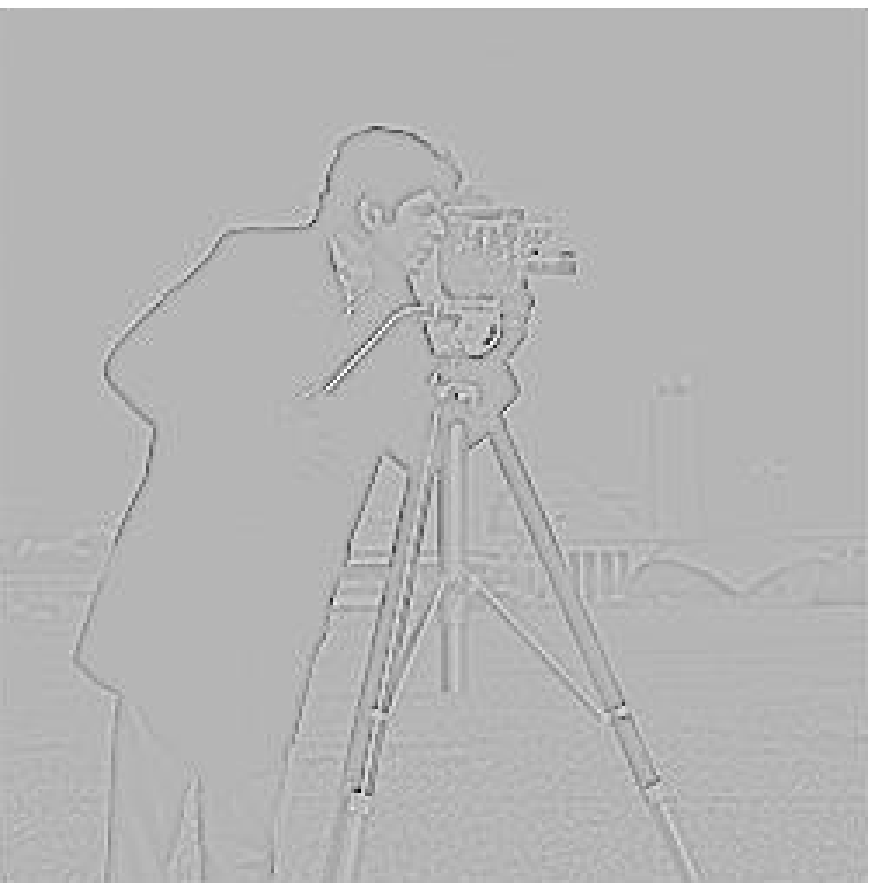}
%
\end{center}
\caption{Computing the Ricci curvature for a pixel: The ``Camera Man'' test
image (top, left), Ricci curvature computed in horizontal (top,
right) and vertical (bottom, left) directions, and the Ricci
(averaged) curvature (bottom, right).}
\end{figure}
As Fig. 5 illustrates, the Combinatorial Ricci curvature we
introduced herein allows even for a non-optimal choice of weights, a very
good approximation of Gaussian curvature of surfaces (i.e. for
gray-scale images). Here, classical Gaussian curvature was computed
using finite element methods standard in Image Processing -- see
\cite{SAZ}. 
%
%
\begin{figure}[h] \label{saucan-fig:LenaK}
\begin{center}
\includegraphics[scale=0.41]{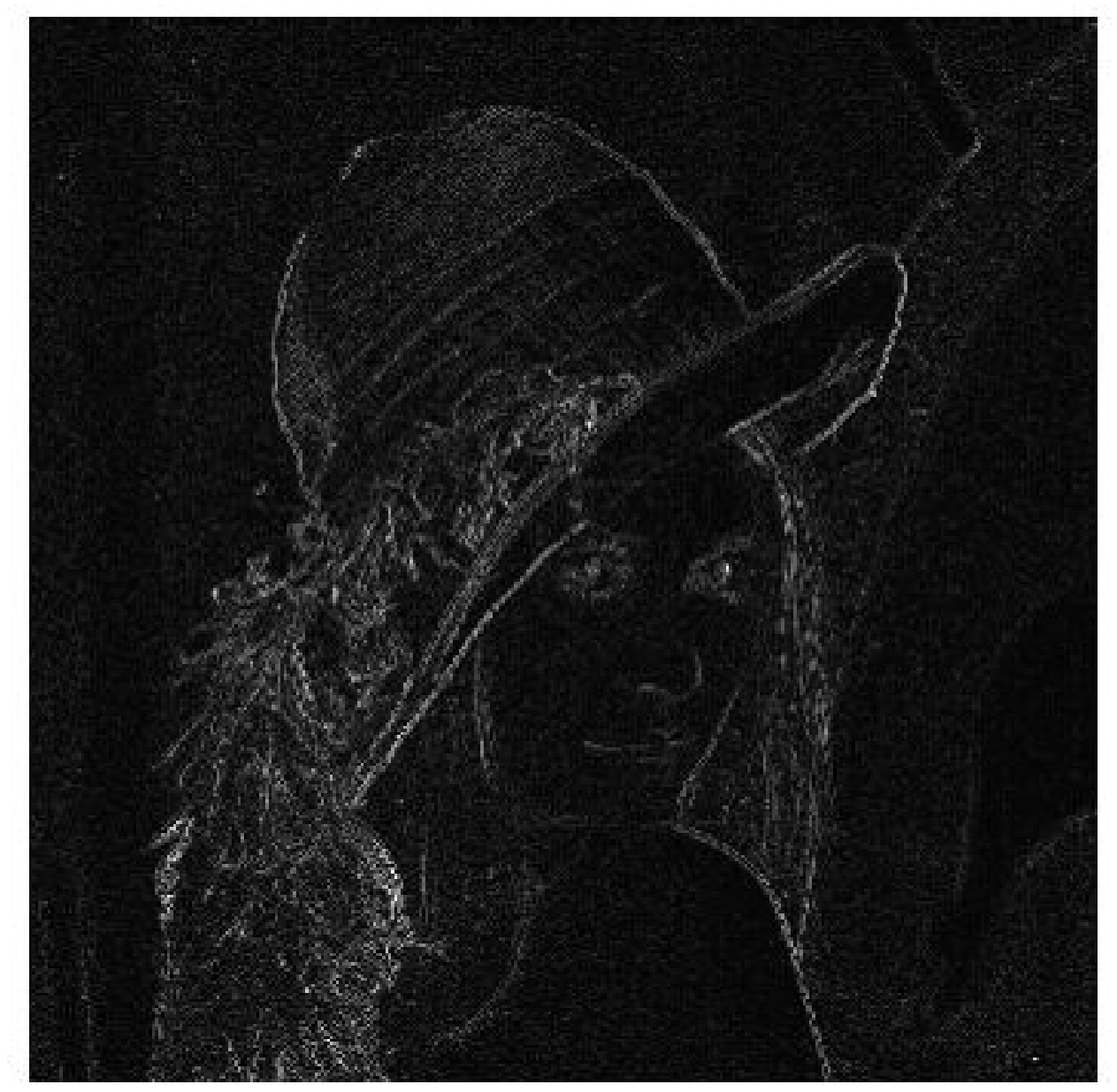}\includegraphics[scale=0.28]{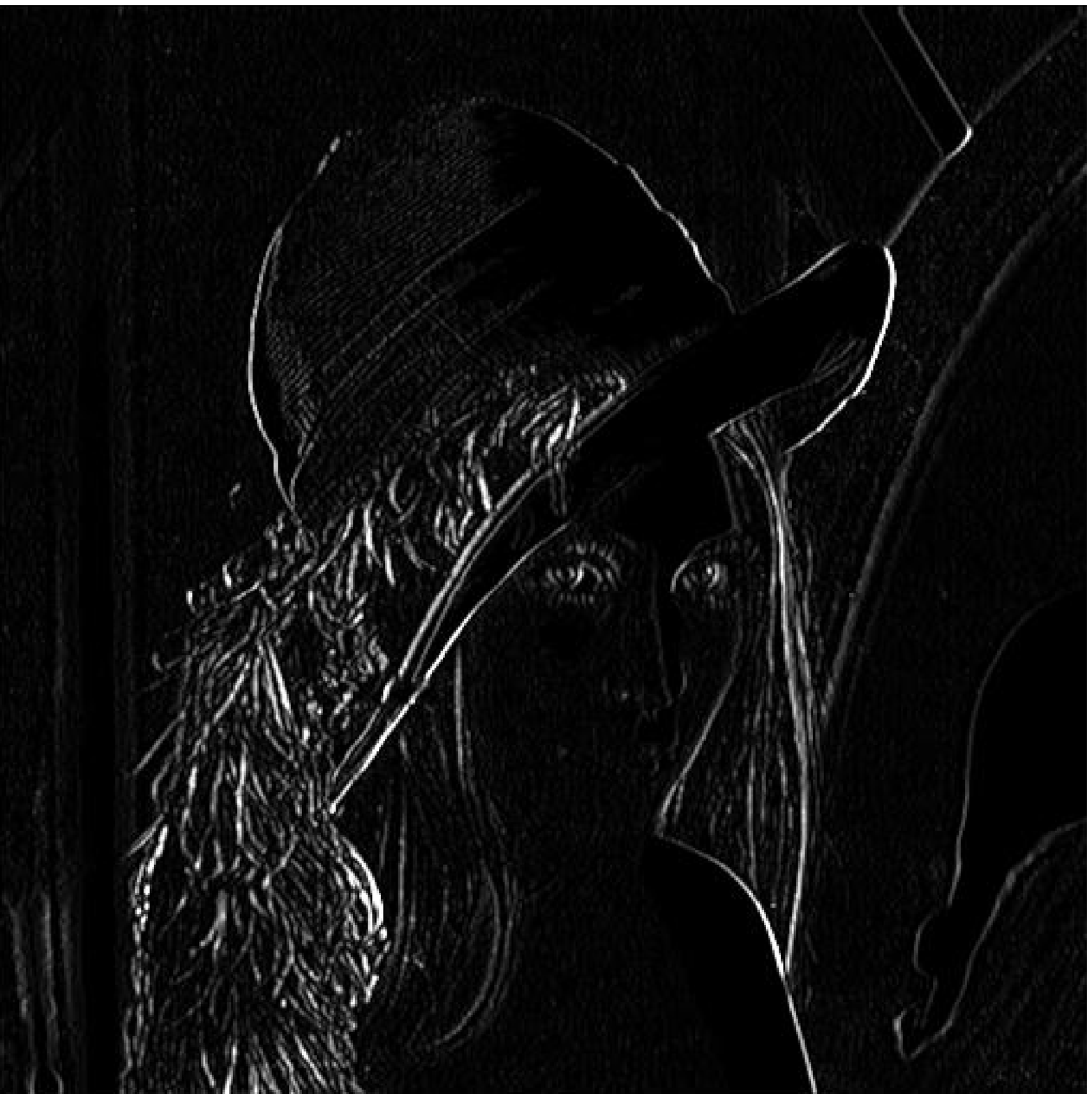}
\end{center}
\caption{Gauss (left) and Combinatorial Ricci (right) comparison. Note the edge detecting capability if the Combinatorial Ricci curvature.}
\end{figure}
In contrast, both the Bochner (and Riemann) Laplacian sharply
diverge from the classical one, e.g. the one obtained by using the
standard {\sf Matlab} function (see Fig. 6). This is not too
surprising, given the fact that 
such a comparison is, in a sense, not relevant, due to the different
dimensionality of the two concepts: The Combinatorial and Bochner
Laplacian are, as stressed above, associated to edges, hence
$1$-dimensional (this being underlined by the notation:
$\Box_1(e_0)$ and $B_1(e_0)$, respectively). In contrast, the
classical Laplacian is a pointwise function, (and, in its discrete
setting, associated to the vertices of a mesh) hence
$0$-dimensional.
However, the Bochner Laplacian proves to be an excellent detector of
``sharp'' edges (see Fig. 7), therefore it may prove to be useful
for contour detection and for segmentation.
\begin{figure}[h] \label{saucan-fig:LenaLap}
\begin{center}
\hspace*{-0.5cm}
\includegraphics[scale=0.46]{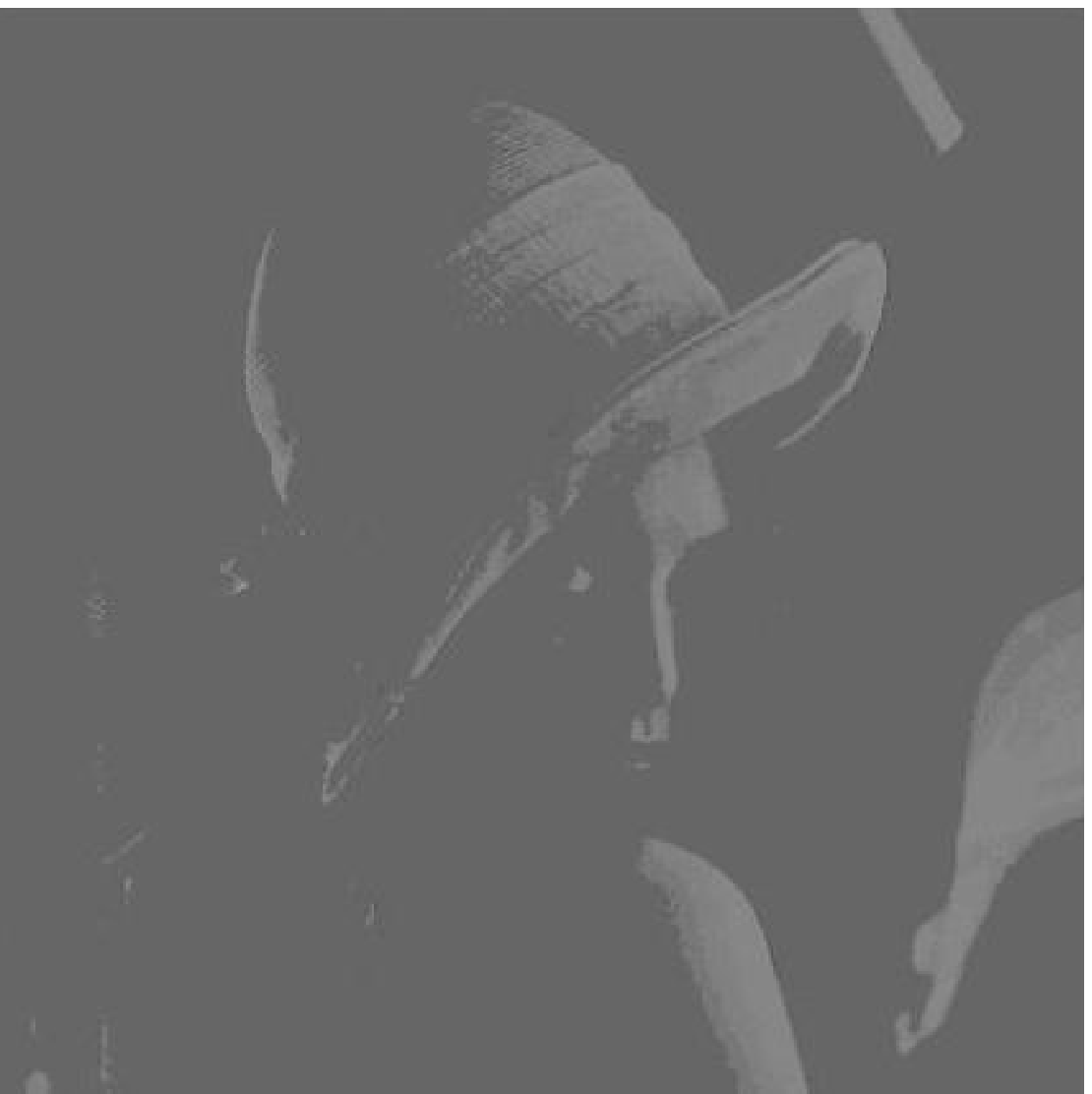}
\includegraphics[scale=0.76]{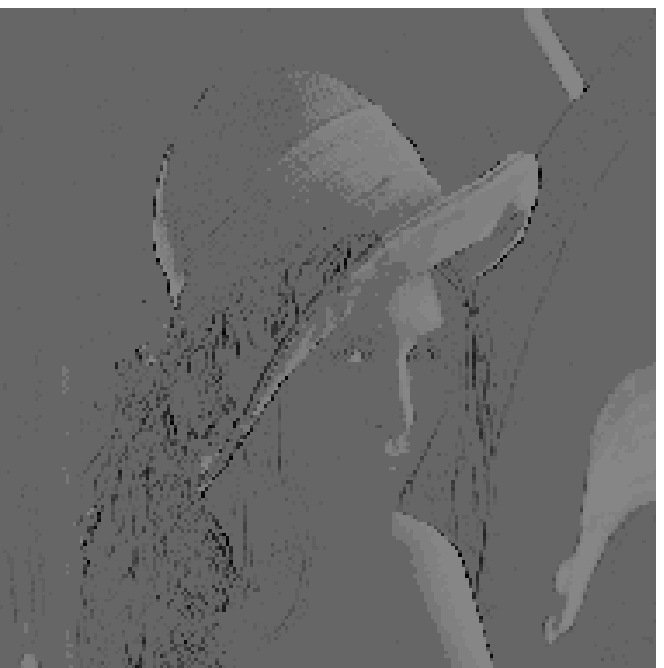}
\end{center}
\caption{Comparison of Different Laplacians: 
The Combinatorial (Forman) Laplacian $\Box_1(e_0)$ (left), and the
Bochner (rough) Laplacian $B_1(e_0)$ (right). Note that the
combinatorial Laplacians  are good detectors of ``sharp'' edges.}
\end{figure}
\begin{figure}[h]  \label{saucan-fig:horizontal}
\begin{center}
\includegraphics[scale=0.66]{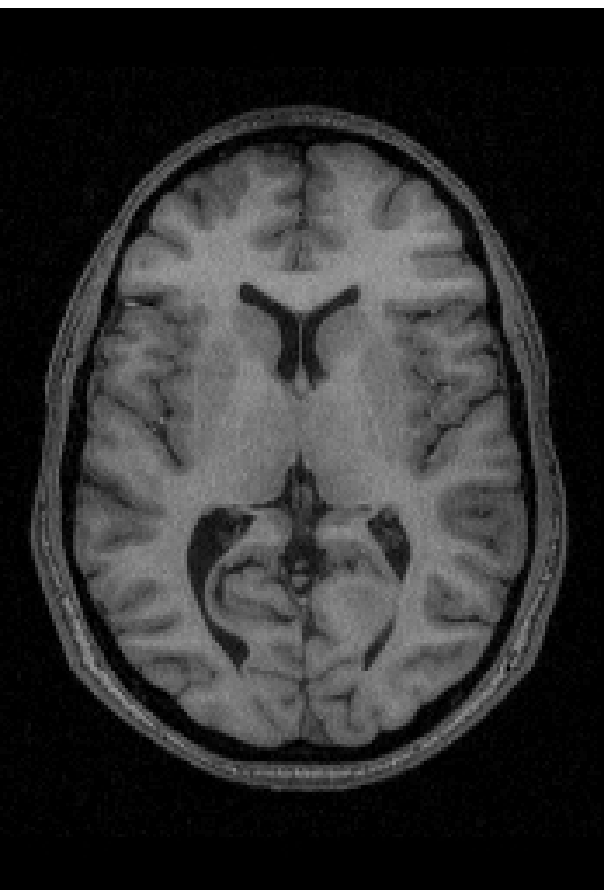}\includegraphics[scale=0.67]{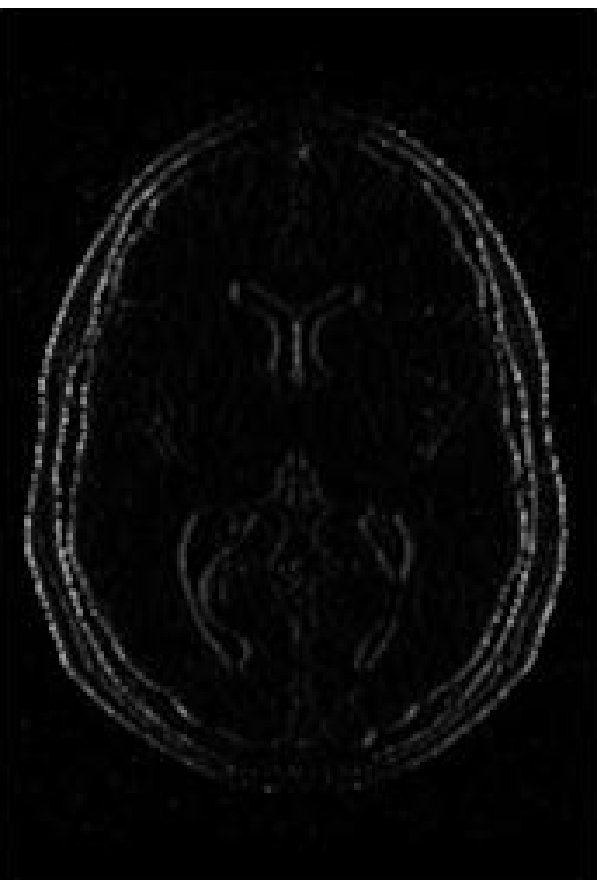}
\hspace*{-0.2cm}\includegraphics[scale=0.69]{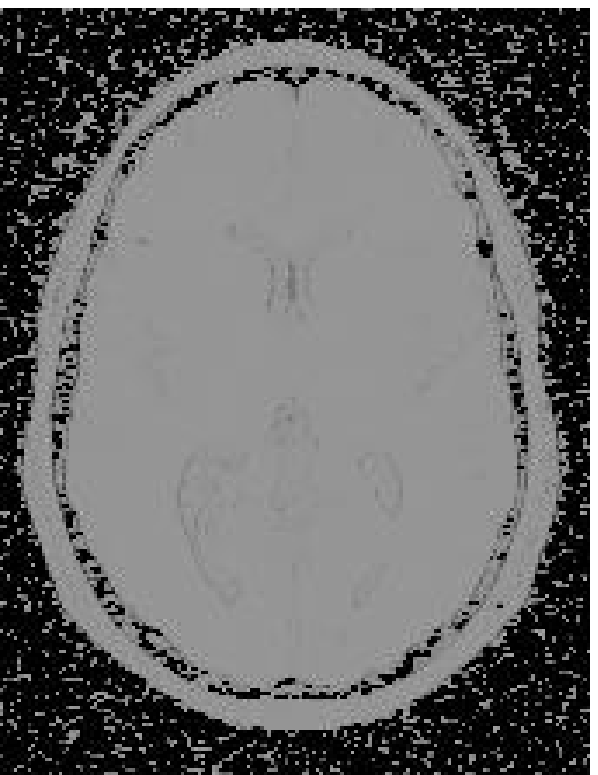}\includegraphics[scale=0.36]{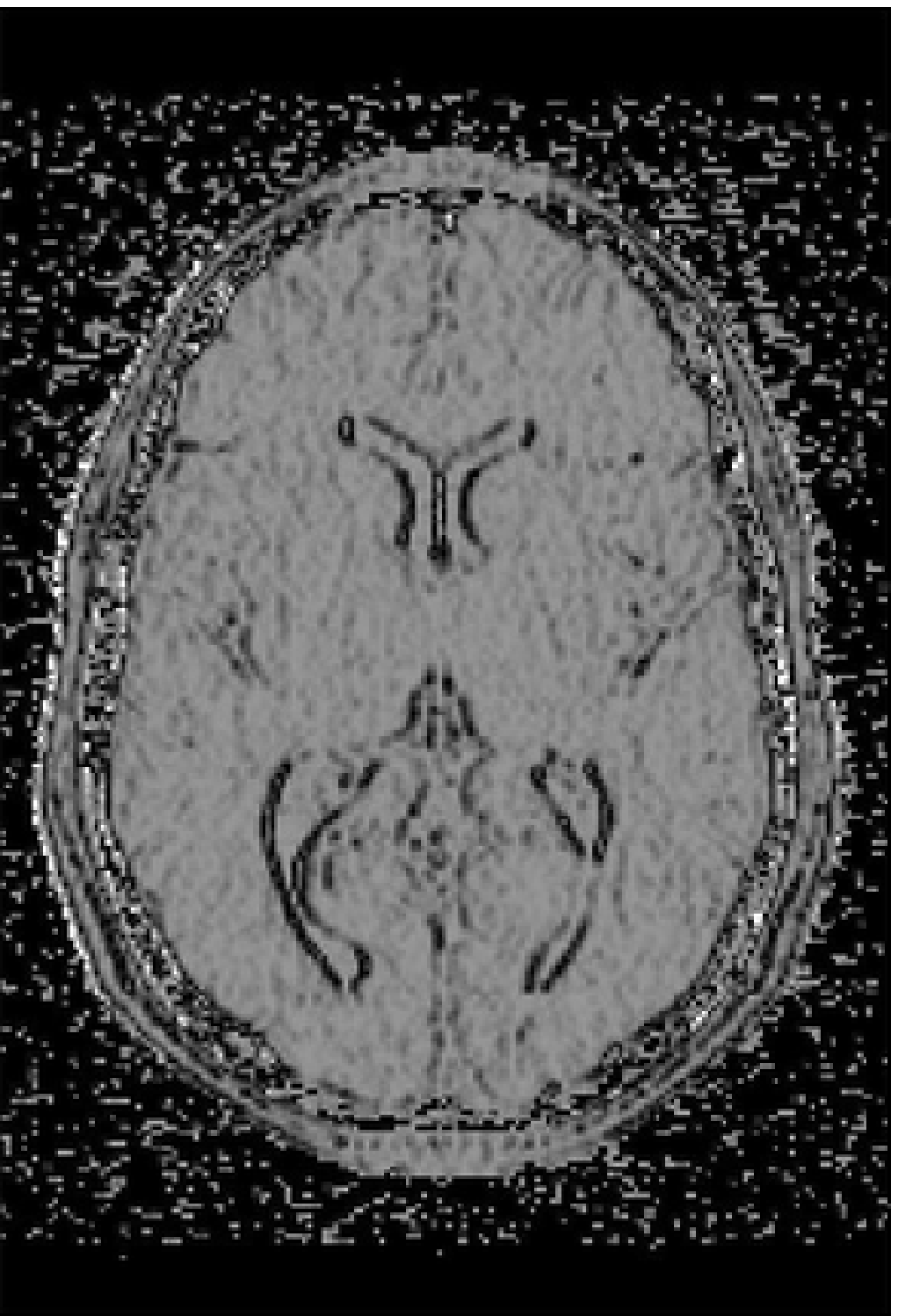}
\end{center}
\caption{Axial brain scan image (top, left), its Combinatorial Ricci
curvature (top, right), the Combinatorial (Forman) Laplacian
$\Box_1(e_0)$ (bottom, left), and the Bochner (rough) Laplacian
$B_1(e_0)$ (bottom, right). (Recall that $B_1(e_0) = \Box_1(e_0) -
{\rm Ric}(e_0)$.)}
\end{figure}
A ``combinatorial diffusion'' was also performed, either by
upsampling, that is by division of the squares (pixels) into
sub-squares of length $1/2$ or $1/3$, or by downsampling, i.e. by
the fusion of 4 or 9 squares into larger ones. However, using the
fusion of 9 squares produces more inferior results.) The weight after such  upsampling would be, for instance $h_\alpha/4$. 
The effect of this last process on $\Box_1(e_0)$ and on $B_1(e_0)$
can be seen in Fig. 8.
\begin{figure}[h]  \label{saucan-fig:LenaUpsampling}
\begin{center}
\includegraphics[scale=0.75]{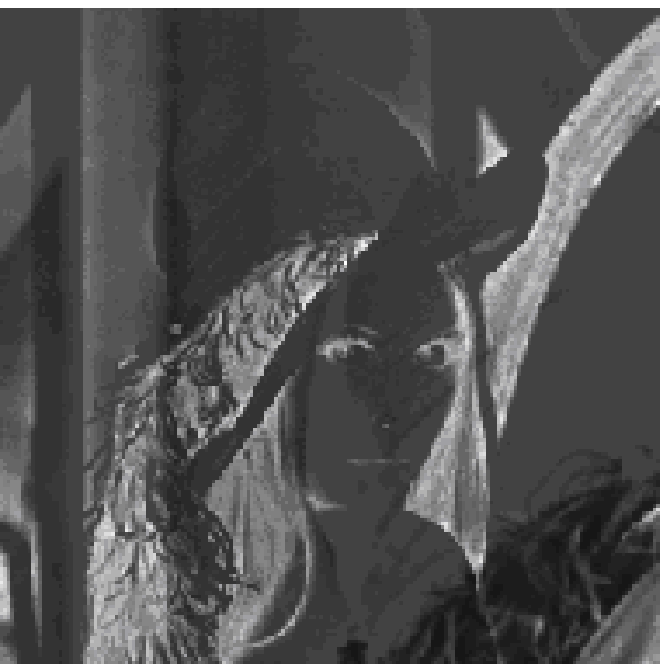}
\includegraphics[scale=0.75]{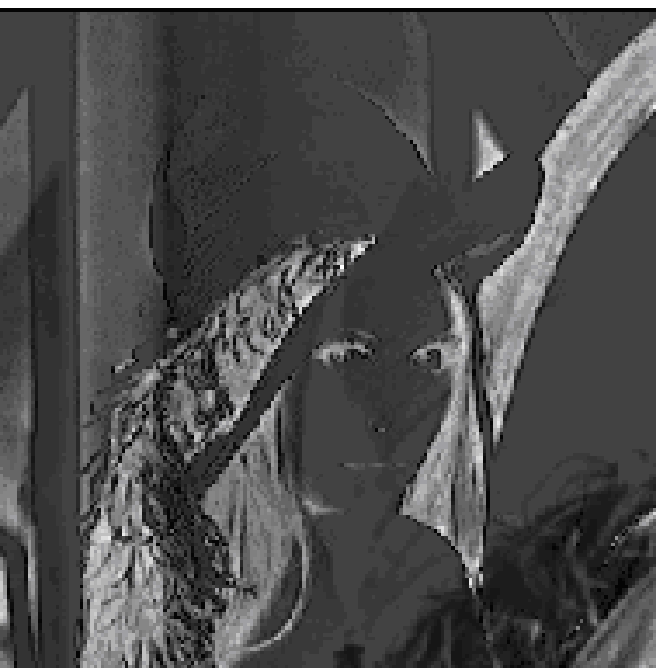}
\end{center}
\caption{Comparison of Different Laplacians after upsampling, by fusion of 4 adjacent squares: From left to right: 
The Combinatorial (Forman) Laplacian $\Box_1(e_0)$ (left), and the
Bochner (rough) Laplacian $B_1(e_0)$ (right).}
\end{figure}

\section{Future Work} \label{saucan-sec:Plans}
We briefly discuss below some of the natural and/or seemingly required directions of further study:
\begin{enumerate}
\item
Evidently, the first task in testing the efficiency of the combinatorial Ricci curvature and Laplacian in medical
imaging is to experiment with voxels, that is to apply the apparatus introduced herein to the analysis of
volumetric data. Such 3- (and even 4-) dimensional manifolds and their evolution in time is most relevant, for
instance, in the analysis of cardiac MRI. This brings us to the following point:

\item
As already mentioned in the introduction, the full power of the Ricci curvature is revealed in the general
heat-type diffusion and in discrete versions of the Ricci flow (and other related flows), that were introduced and
experimented with elsewhere (\cite{CL}, \cite{Gl}, \cite{Gu}, \cite{FL})).
It is only natural to strive to develop and experiment with a
discrete version of the Ricci flow corresponding to the
combinatorial Ricci curvature introduced herein. The Ricci flow is
given by

\[\frac{\partial g}{\partial t} = -2{\rm
Ric}(g(t))\,,\] %
where $g$ denotes the metric of the manifold. (see e.g. \cite{MT}).
Various discretizations are possible for this flow. However, it
seems that given the fact that the Combinatorial Ricci curvature is
an edge measure, the most natural discretization in our context is
the following one, inspired by \cite{FL} (see also \cite{Gu}):

\[\frac{\partial l_{ij}}{\partial t} = -2{\rm
Ric}(e_{ij})\,,\]
where $l_{ij}$ denotes the length of the edge $e_{ij}$.
Experimental work on this type of flow is in progress \cite{us}.

\item As already noted, while we prefer, both for theoretical as well as for practical reasons, to work with
standard weights, Forman's combinatorial version of Ricci curvature
is extremely versatile. Even if restricting oneself to using
standard weights, i.e. proportional to the $p$-dimensional geometric
content ($p$-volume), one still has freedom in choosing the weights
$w_2$ and especially $w_1$ (see Section 2). Hence, to obtain best
result, one can experiment in order to empirically determine, by
using, e.g. variational methods, the optimal standard weights for a
given application of the method.

\item The Combinatorial Laplacian is closely connected (by its very
definition) to the cohomology groups of the cellular complex on
which it operates (see \cite{Fo}). It is natural, therefore,
to apply the results and methods of \cite{Fo} for the estimation of
the dimension, and in some cases even the computation, of the
cohomology groups  (and by duality, of homology groups) of images.
(See \cite{KMM} on this direction of study in Image Processing.)

\end{enumerate}

\subsection*{Acknowledgment} \label{saucan-sec:Ack}
We would like to thank all our students who helped producing many of the images herein.
The first author also wishes to express his gratitude to Professor
Shahar Mendelson -- his warm support is gratefully acknowledged.
The research has been supported in part by Ollendorff Minerva Center for vision and Image Processing, Technion - Israel Institute of Technology.




\begin{thebibliography}{99}


\bibitem{Be1}
Berger, M. {\it Encounter with a Geometer, Part II}, Notices of the
AMS {\bf 47}(3), 326-340, 2000.

\bibitem{Be}
Berger, M. {\em A Panoramic View of Riemannian Geometry}, Springer-Verlag, Berlin, 2003.

\bibitem{CL}
Chow, B. and Luo, F. {\it Combinatorial Ricci Flows on surfaces}, J.
Differential Geometry {\bf 63}, 97-129, 2003.

\bibitem{DLYG}
Dai J., Luo W., Yau S.-T. and Gu X. {\it Geometric accuracy analysis
for discrete surface approximation}, Geometric Modeling and
Processing, 59–72, 2006.

\bibitem{Fo}
Forman, R. {\it Bochner's Method for Cell Complexes and Combinatorial Ricci Curvature}, Discrete and Computational
Geometry, {\bf 29}(3), 323-374, 2003.

\bibitem{Gl}
Glickenstein, D. {\it A combinatorial Yamabe flow in three
dimensions}, Topology {\bf 44}, 791-808, 2005.

\bibitem{H}
Herman, G. T. {\it Geometry of Digital Spaces}, Birkh\"{a}user, 1996.

\bibitem{Gu}
Jin, M., Kim, J. and Gu, X. {\it Discrete Surface Ricci Flow: Theory
and Applications}, Lecture Notes in Computer Science: Mathematics of
Surfaces, {\bf 4647}, 209-232, 2007.

\bibitem{KMM}
Kaczynski, T., Mischaikow, K. and  Mrozek, M. {\it Computational
homology}, Springer-Verlag, New York, 2004.

\bibitem{LL}
Leibon, G. and Letscher, D. {\it Delaunay Triangulations and Voronoi
Diagrams for Riemannian Manifolds}, Proceedings of the Sixteenth
Annual Symposium on Computational Geometry, 341 - 349, 2000.


\bibitem{FL}
Luo, F. {\it A combinatorial curvature flow for compact 3-manifolds
with boundary}, {\it Electronic Research Announcemets of the AMS},
{\bf 11}, 12-20, 2005.

\bibitem{MT}
Morgan, J. and Tian, G. {\it Ricci flow and the Poincare conjecture}, Clay Mathematics Monographs {\bf 3},
Providence, R.I., 2007.

\bibitem{RS}
Rourke, C. P. and Sanderson,  B.J. {\it Introduction to piecewise-linear topology}, Springer-Verlag, Berlin, 1972.

\bibitem{miccai}
Saucan, E., Appleboim, E., Wolansky, G and Zeevi, Y. Y. {\it Combinatorial Ricci Curvature for Image Processing}, Midas Journal, (http://hdl.handle.net/10380/1500), 2008.

\bibitem{us}
Saucan, E., Appleboim, E., Wolansky, G and Zeevi, Y. Y. {\it Combinatorial Ricci Flow for Image Processing},
in preparation.

\bibitem{SAZ}
Saucan, E., Appleboim, E., and Zeevi, Y. Y. {\it Sampling and
Reconstruction of Surfaces and Higher Dimensional Manifolds},
Journal of Mathematical Imaging and Vision, {\bf 30}(1), 105-123,
2008.

\bibitem{YL}
Yakoya, N. and Levine, M. {\it Range image segmentation based on differential geometry: A hybrid
  approach}, IEEE Transactions on Pattern Analysis and Machine
  Intelligence,  {\bf 11}(6), 643-649, 1989.


\end{thebibliography}
\end{document}